\documentclass[10pt,twocolumn]{article}

\usepackage[margin=0.75in]{geometry}
\usepackage{graphicx}
\usepackage{booktabs}
\usepackage{hyperref}
\usepackage{amsmath}
\usepackage{caption}
\usepackage{xcolor}
\usepackage{float}

\hypersetup{colorlinks=true, linkcolor=blue, urlcolor=blue, citecolor=blue}

\title{\textbf{Real2Sim2Real for Vision-Language-Action Manipulation:\\An AMD ROCm-Based Pipeline}}
\author{
Qing Yang$^{1,2}$ \quad Xun Wang$^{1}$ \quad Ziguan Wang$^{1}$\\
Zhenjiang Li$^{1}$ \quad Hongqiang Wang$^{1}$ \quad Dongdong Weng$^{2}$\\[4pt]
\small $^{1}$AMD AIG Team \qquad $^{2}$Beijing Institute of Technology\\
\small \textit{Spatial Interaction and Embodied Intelligence Forum, WAIC Academic (WAICA) 2026}
}
\date{}

\begin{document}
\maketitle

\begin{abstract}
Physical AI -- the integration of large vision-language-action (VLA) models with embodied agents that act in the real world -- has emerged as the next major frontier for AI, echoed by industry leaders such as Jensen Huang (``the next big thing is Physical AI, AI with a body,'' GTC Paris, June 2025) and Dr. Lisa Su (``we're entering the world of Physical AI ... this is where AI enters the real world,'' CES 2026). This paper presents an end-to-end, fully AMD-accelerated technology stack for embodied manipulation, spanning data-center training silicon, Radeon PRO simulation/rendering GPUs, and Ryzen AI edge compute, unified by the open ROCm software stack. We demonstrate that training and deploying VLA-based manipulation policies does not require a CUDA-locked ecosystem. Four progressive demonstrations are presented: (1) a Sim-to-Real manipulation pipeline trained with SmolVLA and deployed on a physical Franka arm; (2) a semantic, language-grounded object-selection task (``one-of-three''); (3) a Real2Sim synthetic-data generation pipeline that fuses 3D Gaussian Splatting (3DGS) reconstructions of real scenes with the Genesis physics engine; and (4) large-scale reinforcement learning for quadruped and humanoid locomotion benchmarked across multiple hardware platforms. All pipelines run natively on ROCm + PyTorch on RDNA4 (Radeon AI PRO R9700) and RDNA3.5 (Radeon PRO W7900) hardware and are reproducible on the free Radeon Cloud Platform.
\end{abstract}

\section{Introduction}

The trajectory of modern AI has moved from Generative AI, to Agentic AI, and now to \emph{Physical AI}: systems that perceive, reason, and act inside the physical world rather than a purely digital context. Two data points frame this shift. At GTC Paris in June 2025, Jensen Huang stated that ``the next big thing is Physical AI, AI with a body.'' At CES 2026 in Las Vegas, AMD's own Dr. Lisa Su reinforced the point: ``Now we're entering the world of Physical AI. This is where AI enters the real world.''

Physical AI raises the computational bar considerably relative to prior AI waves: it requires (i) massive parallel compute for photorealistic and physically-accurate simulation, (ii) large-memory accelerators for training vision-language-action (VLA) policies, and (iii) low-latency, memory-efficient inference at the edge, directly on or near the robot. AMD's hardware portfolio is well-positioned to cover this entire pipeline end to end:

\begin{itemize}
\item \textbf{Data Center Series} -- Gen 1 (192GB HBM3, 5.3TB/s) is capable of training a full VLA policy on a single GPU; Gen 2 (288GB HBM3E, 8TB/s, CDNA4) scales simulation rollouts for larger fleets and longer horizons.
\item \textbf{Radeon PRO Series} (simulation and rendering) -- the W7900 (48GB GDDR6, 864GB/s) targets high-fidelity physics and rendering simulation, while the R9700 PRO (32GB GDDR6, 640GB/s, 128 AI accelerators) enables desktop-local AI and simulation workloads.
\item \textbf{Ryzen AI Series} (edge / on-robot) -- a unified memory architecture (UMA) lets the CPU, GPU, and NPU share a single memory pool with no copy overhead; the Max+ 395 ships with 128GB today, rising to 192GB with the next-generation Gorgon Halo silicon in 2026.
\end{itemize}

\begin{figure*}[ht]
\centering
\includegraphics[width=\linewidth]{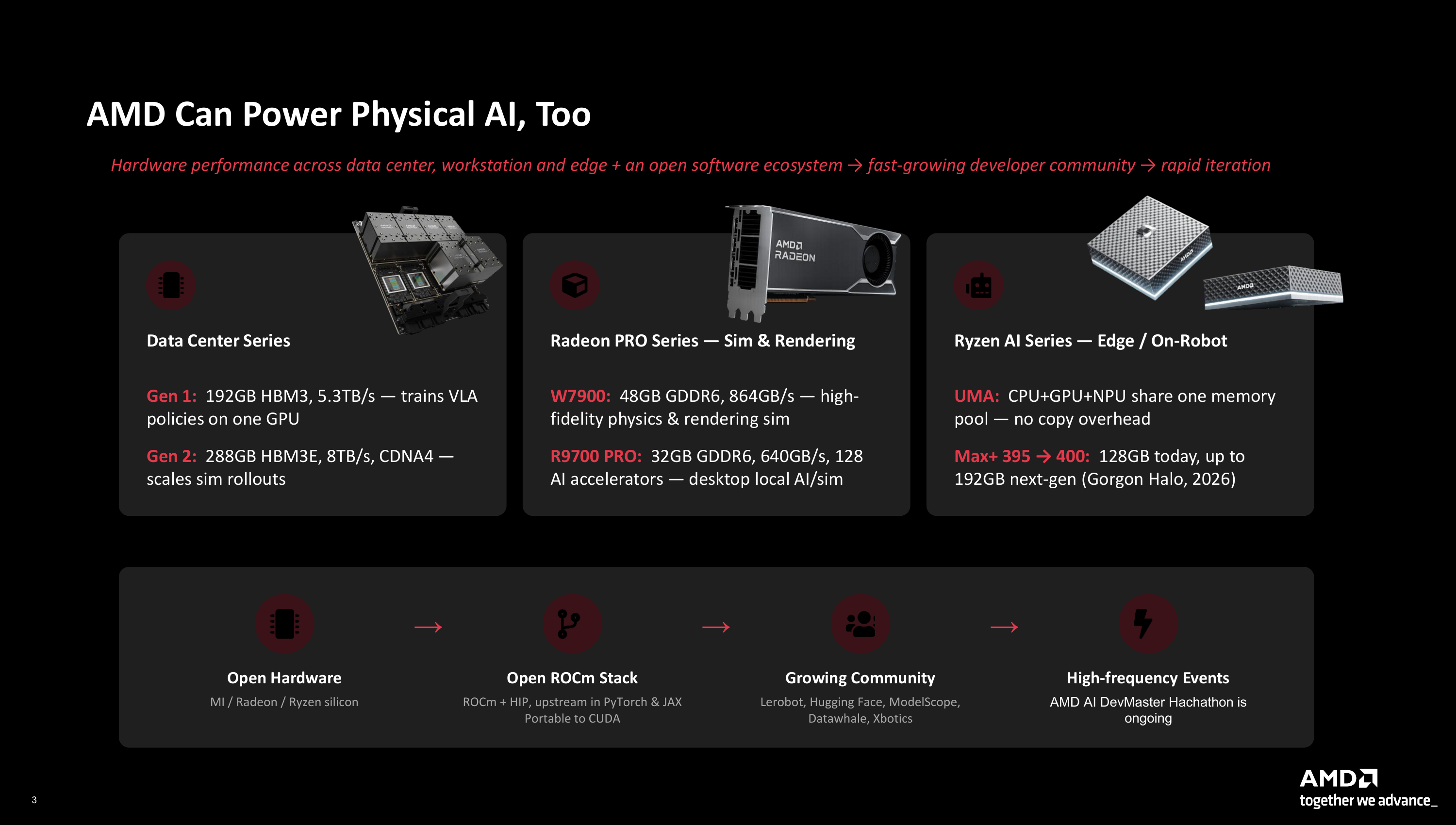}
\caption{AMD's hardware-to-community stack for Physical AI: open silicon (MI/Radeon/Ryzen) $\rightarrow$ the open ROCm + HIP software stack (upstreamed into PyTorch and JAX, portable from CUDA) $\rightarrow$ a growing open-source community (LeRobot, Hugging Face, ModelScope) $\rightarrow$ high-frequency community events such as the ongoing AMD AI DevMaster Hackathon.}
\label{fig:hwstack}
\end{figure*}

Underpinning all three tiers is the open ROCm~\cite{rocm} + HIP software stack, which is upstreamed directly into PyTorch and JAX and is portable from existing CUDA codebases, and an increasingly active open-source robotics community built around LeRobot~\cite{lerobot}, Hugging Face, and ModelScope (Figure~\ref{fig:hwstack}). The combination of open hardware, an open compute stack, and a fast-growing developer community enables rapid iteration -- exactly the property Physical AI research needs, since each simulation-training-deployment cycle must be repeated many times to close the sim-to-real gap.

This paper documents four technology demonstrations built entirely on this AMD stack and presented at the Embodied Intelligence \& Spatial Interaction Forum of WAIC Academic (WAICA) 2026~\cite{waica}: (1) Sim-to-Real manipulation, (2) semantic ``one-of-three'' object selection, (3) a Real2Sim synthetic data-generation pipeline, and (4) large-scale reinforcement learning for legged and humanoid locomotion. Each demo is analyzed both functionally (what capability it demonstrates) and quantitatively (what it costs / how fast it runs on AMD hardware), and every pipeline is reproducible free of charge through the Radeon Cloud Platform.

\section{Demo 1: Sim-to-Real Manipulation~\cite{sim2real}}
\label{sec:demo1}

\begin{figure}[]
\centering
\includegraphics[width=\linewidth]{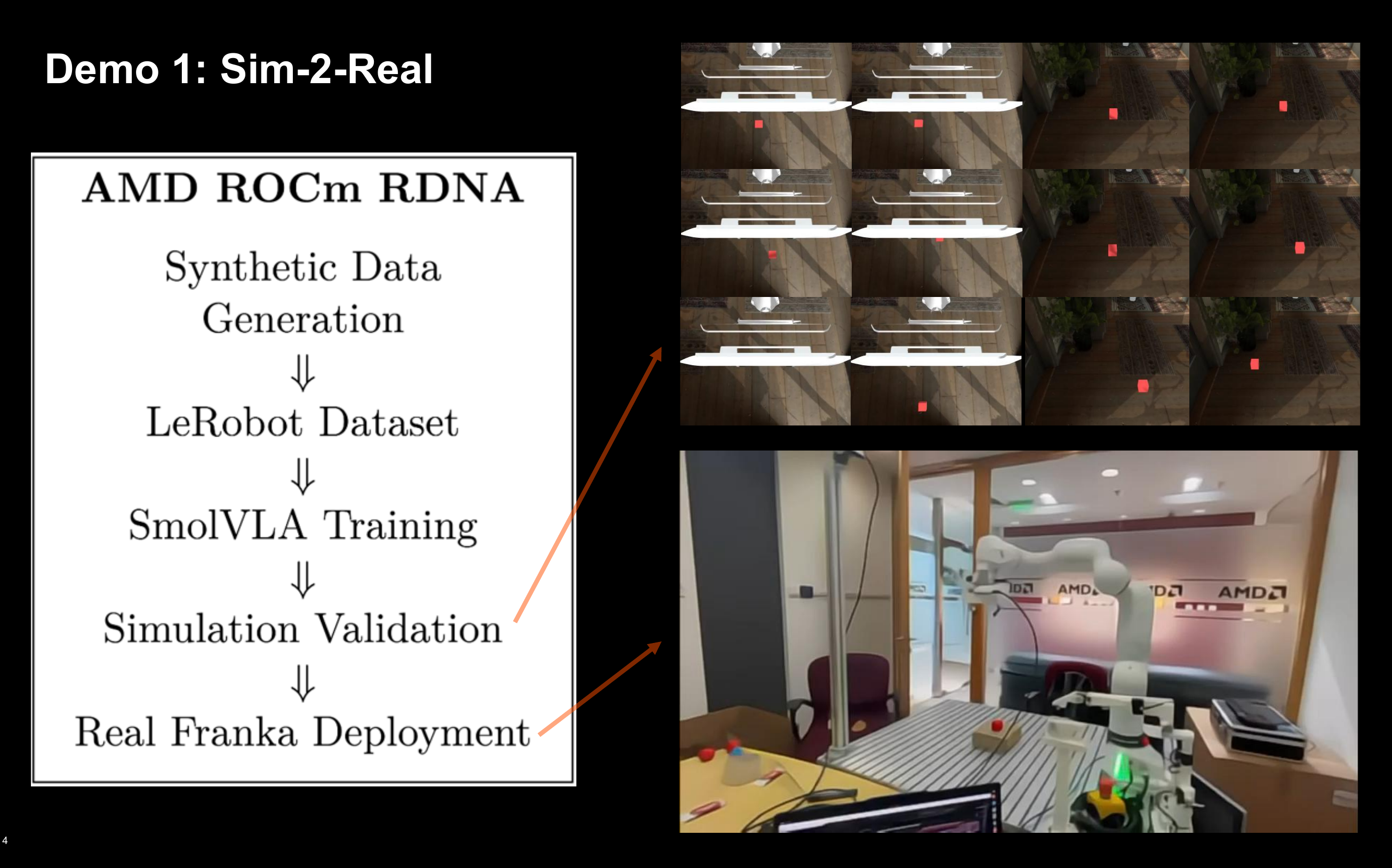}
\caption{Sim-to-Real pipeline: synthetic data generation in Genesis $\rightarrow$ LeRobot dataset $\rightarrow$ SmolVLA training $\rightarrow$ simulation validation $\rightarrow$ deployment on a real Franka arm.}
\label{fig:pipeline}
\end{figure}

This demo showcases an end-to-end Vision-Language-Action (VLA) workflow accelerated by AMD ROCm and Radeon GPUs, covering robot simulation, policy training, and real-world deployment. Robot demonstrations are generated in the Genesis physics simulator~\cite{genesis}, converted into the LeRobot dataset format~\cite{lerobot}, and used to train a SmolVLA~\cite{smolvla} policy entirely on the AMD platform. The trained policy is then deployed to a physical Franka~\cite{franka} robotic arm for closed-loop manipulation. This solution was first brought up in AI DevDay work by David Li~\cite{https://github.com/PhysicalAI-AIM/Robot_synthetic_data_generation_workshop}.

Figure~\ref{fig:pipeline} illustrates the complete workflow. The accompanying demonstration includes three video clips: \textit{00-sim-grasping.mp4}, showing autonomous object grasping in the Genesis simulator; \textit{01-real-grasping.mp4}, demonstrating sim-to-real object grasping on a Franka robotic arm using the same trained policy; and \textit{02-sim-place\_banana\_into\_bowl.mp4}, presenting a continuous pick-and-place task in simulation. Together, the figure and videos demonstrate AMD ROCm and Radeon GPUs as a unified platform for embodied AI simulation, VLA training, and robotic deployment.

\paragraph{Analysis.} The pipeline was validated on both RDNA4 (Radeon AI PRO R9700) and RDNA3.5 (Radeon PRO W7900) hardware, running natively on ROCm + PyTorch without any CUDA translation layer. Synthetic-data generation achieved a 100\% success rate for the grasping task distribution used to seed training. Fine-tuning the SmolVLA-450M backbone converged in approximately 7--11 minutes for 4{,}000 training steps, with peak VRAM usage under 2.4GB -- low enough that policy training, simulation rollout, and validation can share a single desktop-class Radeon PRO GPU rather than requiring a dedicated training cluster. The full loop, from synthetic demonstration generation through simulation validation to a working real-Franka deployment, completes in under one hour, which is the key enabler for the rapid iterate-and-validate cycle Physical AI research depends on. Qualitatively, the policy trained purely on simulated demonstrations transferred directly to the physical arm without additional real-world fine-tuning, indicating that even the comparatively simple procedural randomization used to seed this Genesis-based data generation is already sufficient to close a meaningful part of the sim-to-real gap for basic pick-and-place tasks. This result motivates the more systematic, real-scene-grounded randomization pipeline explored in Demo 3 (Section~\ref{sec:demo3}), which targets harder, more visually diverse manipulation settings.

\section{Demo 2: One-of-Three, Semantic Select}
\label{sec:demo2}

\begin{figure}[h]
\centering
\includegraphics[width=\linewidth]{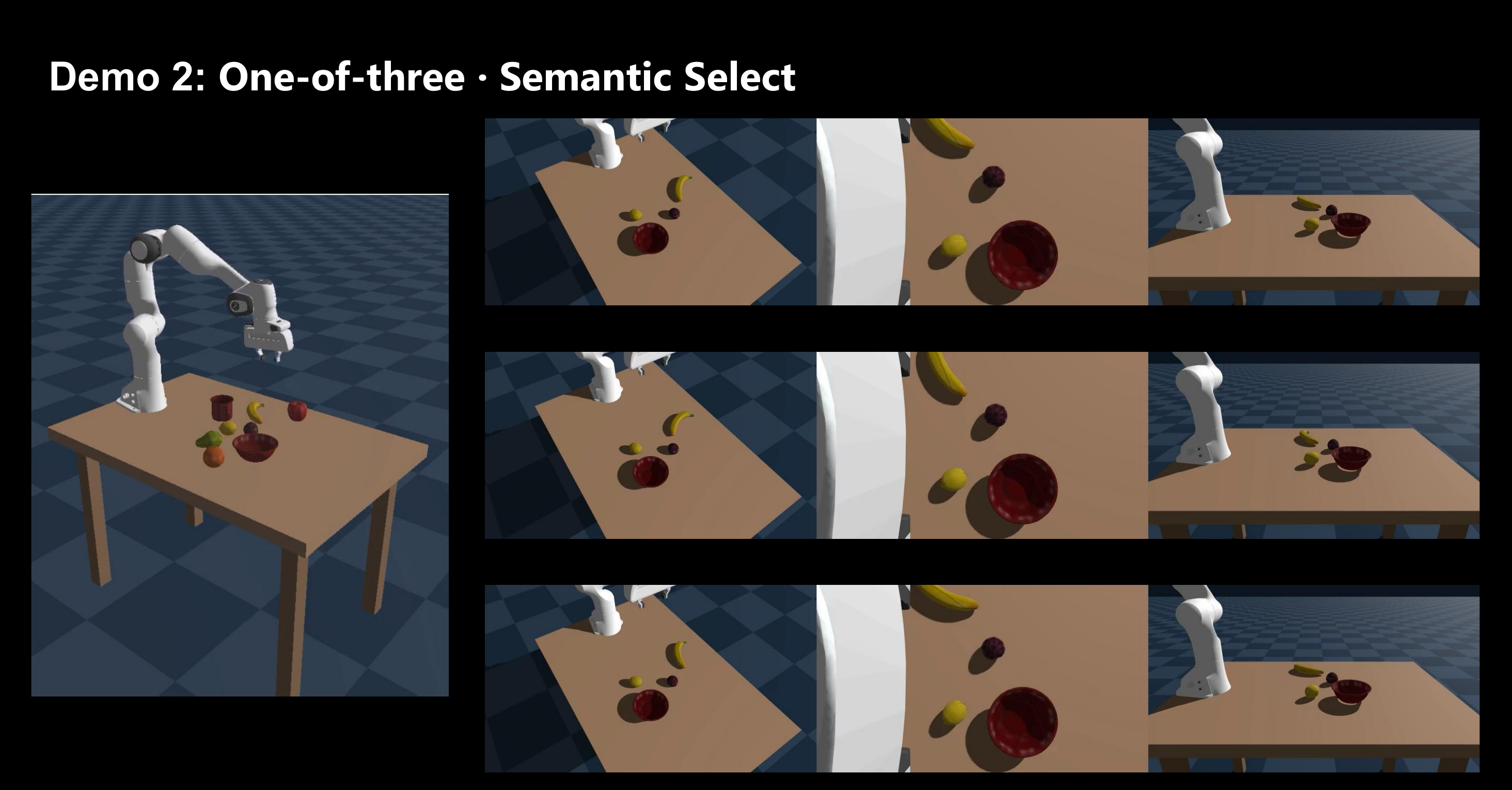}
\caption{Semantic object selection: given a natural-language instruction, the policy must identify and place the correct target object (e.g., banana, lemon, apple) among visually similar distractors on the table.}
\label{fig:semantic}
\end{figure}

Building on Demo 1's grasping capability, Demo 2 evaluates whether the same VLA policy family can ground natural-language instructions in cluttered scenes. A Franka arm is presented with three or more candidate objects on a table (e.g., a banana, grapes, and a bowl of fruit) and must select and place the specific object named in the instruction while ignoring the distractors (Figure~\ref{fig:semantic}).

\paragraph{Analysis.} Unlike Demo 1, which requires only motor competence, Demo 2 stresses the language-conditioning branch of the VLA policy: the model must fuse the language embedding of the instruction with the visual scene to resolve reference ambiguity among near-identical object categories (e.g., choosing the banana rather than the grapes when both are visible in the same clutter of fruit). Because the underlying SmolVLA backbone and ROCm training pipeline are shared with Demo 1, this task adds essentially no additional infrastructure cost -- the same 450M-parameter checkpoint and sub-2.4GB VRAM training footprint apply -- while validating semantic grounding as a separate axis of policy quality from raw manipulation accuracy. This separation of concerns (motor competence vs.\ semantic grounding) is useful for diagnosing failure modes: a policy that grasps well but selects the wrong object indicates a language-vision fusion deficiency rather than a control-policy deficiency, and vice versa.

\section{Demo 3: Real2Sim Synthetic Training Data Pipelines}
\label{sec:demo3}

\begin{figure}[h]
\centering
\includegraphics[width=\linewidth]{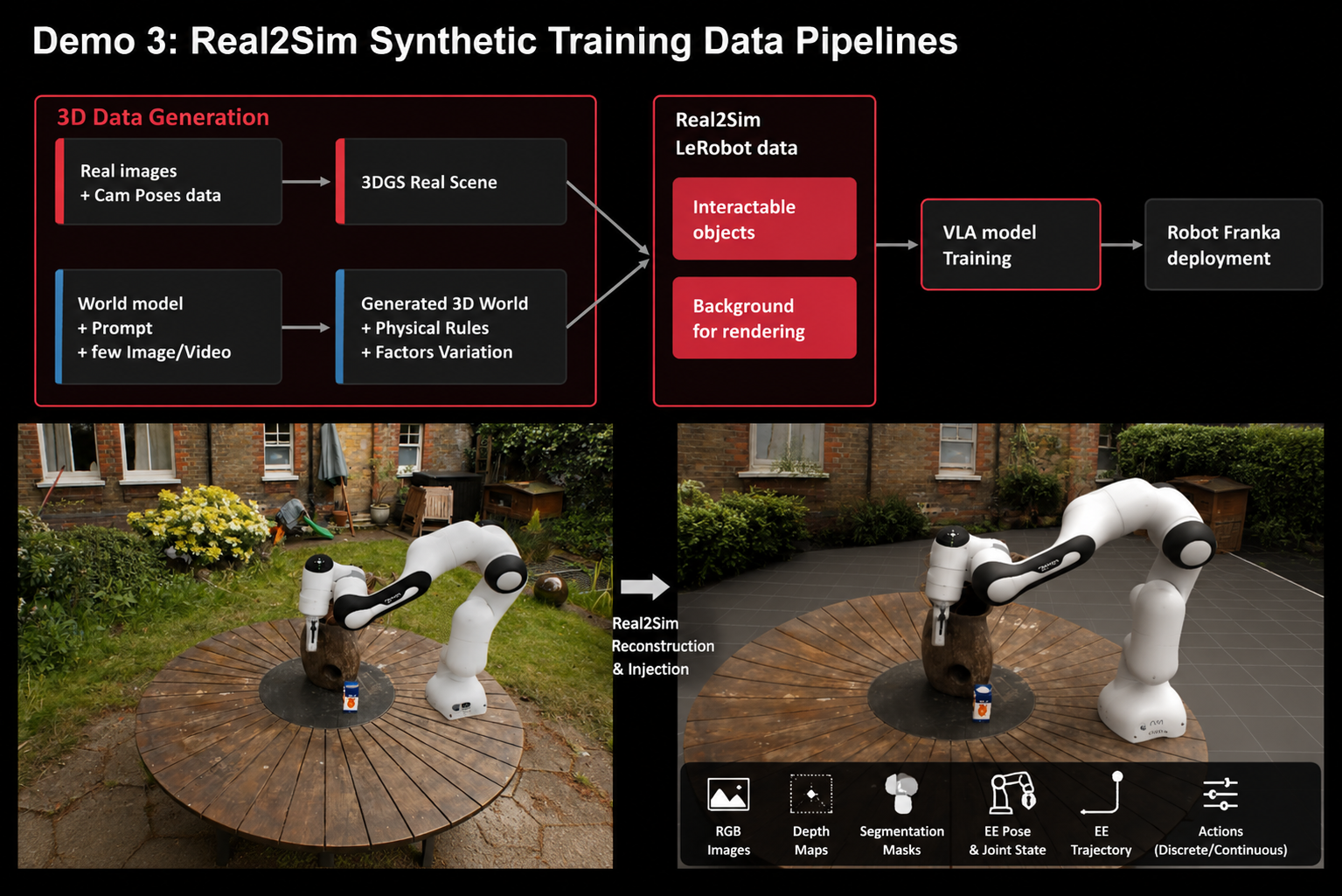}
\caption{Real2Sim pipeline: real images and camera poses are reconstructed into a 3D Gaussian Splatting (3DGS) scene, fused with a world-model-generated physics-aware environment, and converted into interactive LeRobot-format training data for VLA policy training and Franka deployment.}
\label{fig:real2sim}
\end{figure}

Demo 3 addresses the pipeline in the opposite direction from Demo 1: instead of generating data purely in simulation, it reconstructs \emph{real} environments and injects them back into simulation so that photorealistic, physically-interactive training data can be produced at scale (Figure~\ref{fig:real2sim}). Built entirely on the AMD AI platform, this Real2Sim pipeline combines 3D Gaussian Splatting (3DGS)~\cite{gsplat} reconstruction with the Genesis~\cite{genesis} physics engine to automate the generation of robot training data directly from real-world captures. By preserving the geometry, appearance, and spatial layout of real environments while introducing controllable physical interactions, the framework bridges the gap between perception and manipulation, enabling robots to learn from environments that closely resemble their deployment scenarios.

The pipeline proceeds in four stages. First, AMD Radeon GPUs and the ROCm software stack are used to perform high-fidelity 3DGS reconstruction of a real environment, producing a Gaussian scene representation that can be rendered in real time. Second, this Gaussian point cloud is loaded into the Genesis simulator as a photorealistic, static background, into which a Franka Emika Panda arm is introduced as the acting agent alongside interactive objects endowed with physical properties (mass, friction, collision geometry), enabling representative manipulation tasks such as grasping, transporting, placing, and opening/closing articulated objects. Third, Genesis's high-precision physics simulation automatically generates the full set of VLA-relevant training signals for each rollout: end-effector trajectories, multi-view RGB images, depth maps, segmentation masks, end-effector poses, joint states, and discrete/continuous actions, all exported in a format compatible with LeRobot~\cite{lerobot} and other mainstream robot-learning frameworks. In addition, synchronized language annotations and task descriptions can be associated with each trajectory, facilitating the training of multimodal Vision-Language-Action policies that jointly reason over visual observations, language instructions, and robot actions. Fourth, the pipeline exposes randomization controls over object layout, material/texture, lighting, camera viewpoint, robot initial pose, and task goal, enabling large-scale, high-diversity data augmentation while preserving the visual fidelity of the original real-world scene. Such controllable domain randomization substantially increases the diversity of training samples without requiring repeated real-world data collection, significantly reducing annotation effort and improving policy robustness across unseen environments.

The resulting datasets are directly consumable by Vision-Language-Action (VLA) models, manipulation policies, and imitation-learning (IL) pipelines for both training and evaluation, providing a complementary route -- alongside the simulation-first pipeline of Demo~1 -- to narrowing the Sim2Real gap. More broadly, the proposed Real2Sim workflow demonstrates how modern neural scene representations and physics simulation can be seamlessly integrated on AMD AI hardware to build scalable, automated data-generation pipelines for next-generation embodied AI systems. The reconstruction-to-simulation bridge is implemented as an open-source plugin~\cite{vkgsplat}.

\paragraph{Analysis.} The distinguishing property of the Real2Sim pipeline relative to Demo 1's purely synthetic path is that the \emph{background and static geometry} are reconstructed from real sensor data (images + camera poses) via 3DGS, while the \emph{dynamic, interactive} content (the robot arm and manipulable objects) is supplied procedurally with correct physical parameters. This hybrid representation gets the best of both worlds: the visual realism and geometric fidelity of a real captured scene (avoiding the ``simulation look'' that can widen the sim-to-real visual gap), combined with the unlimited, safe, and inexpensive rollout generation that only simulation can provide. Because 3DGS reconstruction, neural rendering, physics simulation, and VLA model training are all compute-bound but heterogeneous in their arithmetic (rasterization/rendering vs.\ rigid-body physics vs.\ transformer training), this pipeline is a good stress test of AMD's unified compute story: the same Radeon PRO GPU (via ROCm) that reconstructs the 3DGS scene and rasterizes it in Genesis also trains the downstream VLA policy, without a hand-off to a separate CUDA-only rendering or training stack. A runnable version of this pipeline is published on the Radeon Cloud Platform (Section~\ref{sec:resources}) so that developers can reproduce the full Real2Sim loop -- from a handful of real photographs to a trained manipulation policy -- without needing physical access to AMD hardware.

\section{Demo 4: RL Training with Unilab on the AMD Platform}
\label{sec:demo4}

\begin{figure}[h]
\centering
\includegraphics[width=\linewidth]{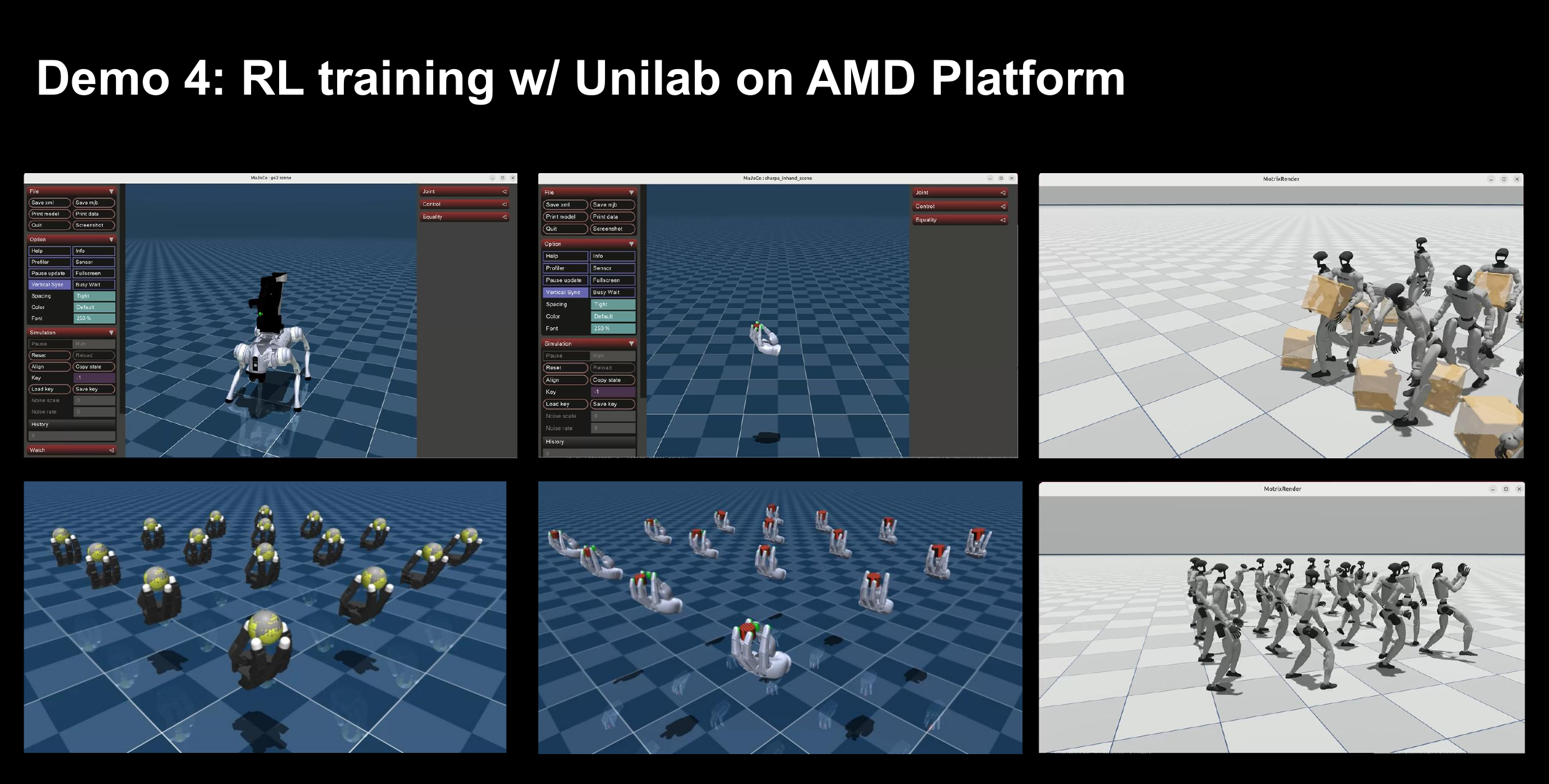}
\caption{Reinforcement-learning training of quadruped (Unitree Go2) and humanoid (Unitree G1) robots using the Unilab framework, including whole-body tracking (WBT), walking, joystick locomotion, and a dynamic flip task.}
\label{fig:rl}
\end{figure}

Beyond manipulation, AMD's platform is also benchmarked on locomotion-focused reinforcement learning (RL) using the Unilab~\cite{unilab} framework. Quadruped (Unitree Go2) and biped/humanoid (Unitree G1) robots are trained in parallel, massively-batched simulation to perform whole-body tracking (WBT), walking, joystick-commanded locomotion, and a dynamic flip maneuver (Figure~\ref{fig:rl}), using a mix of algorithms (FastSAC, FlashSAC, PPO) matched to each task's characteristics.

\begin{table}[]
\centering
\caption{RL training throughput (steps/second, higher is better) across hardware platforms and task/algorithm pairs.}
\label{tab:rl}
\resizebox{\linewidth}{!}{%
\begin{tabular}{lrrrr}
\toprule
\textbf{Device} & \textbf{FastSAC} & \textbf{FastSAC} & \textbf{FlashSAC} & \textbf{PPO}\\
 & \textbf{G1 WBT} & \textbf{G1 Walk} & \textbf{Go2 Joy.} & \textbf{G1 Flip}\\
\midrule
RTX 4090 (baseline) & 58.8 & 18.3 & 6.0 & 109.0 \\
RTX 4090 + AMD 9950X3D & 18.5 & 3.0 & 1.1 & 16.4 \\
AMD 8060S + AI MAX 395 & 33.6 & 9.4 & 4.2 & 19.6 \\
Apple M5 Max & 75.0 & 18.8 & 4.5 & 16.8 \\
\bottomrule
\end{tabular}%
}
\end{table}

\paragraph{Analysis.} Table~\ref{tab:rl} reports steps-per-second throughput for each device/task pair; several observations follow. First, throughput is highly task-dependent rather than purely hardware-dependent: the RTX 4090 baseline leads decisively on the PPO/G1 Flip task (109.0 steps/s) but is the slowest configuration on FlashSAC/Go2 Joy (6.0 steps/s), suggesting that algorithm-specific CPU/GPU synchronization and batch-size sensitivity dominate over raw GPU throughput for some RL algorithms. Second, the ``RTX 4090 + AMD 9950X3D'' configuration -- pairing an NVIDIA discrete GPU with an AMD CPU -- underperforms the pure RTX 4090 baseline across all four tasks, which is consistent with RL training in Unilab being at least partially CPU/host-bound (e.g., simulation stepping, environment resets, or data transfer), so the CPU choice materially affects end-to-end throughput even when the GPU is unchanged. Third, the integrated AMD Ryzen AI MAX 395 (with its 8060S iGPU and unified memory architecture) delivers competitive, and in some cases (FastSAC/G1 WBT, FlashSAC/Go2 Joy) superior, throughput relative to the discrete-GPU-plus-AMD-CPU configuration, despite being a single-chip mobile/edge-class part rather than a discrete data-center or workstation GPU -- evidence that the UMA design's elimination of host-device copy overhead is a meaningful advantage for RL workloads with tight simulate-train loops. Taken together, these results support positioning the AMD AI MAX platform not merely as an edge-inference target but as a viable on-device or workstation-class RL training platform for legged locomotion research, while also highlighting that cross-hardware RL benchmarking must be read per-algorithm rather than as a single aggregate number.

\section{Developer Resources}
\label{sec:resources}

All four pipelines are reproducible free of charge on the Radeon Cloud Platform~\cite{radeoncloud} (\url{https://radeon.anruicloud.com}), which hosts runnable GPU notebooks covering the full breadth of this paper's demos -- including the Real2Sim ``Robot Synthetic Data Generation'' workshop notebook of Section~\ref{sec:demo3} (\url{https://radeon.anruicloud.com/templates/7/preview}), which executes an end-to-end synthetic-data-generation, SmolVLA fine-tuning, and closed-loop evaluation pipeline on RDNA4 (R9700) or RDNA3.5 (W7900) nodes in approximately 30 minutes of wall-clock time, entirely on-device with no external dataset download required.

Developers can also access ROCm tutorials, sample projects, and community support through the AMD Developer Community (\url{https://developer.amd.com.cn/}), the AMD AI Developer Program (China, offering 200 free hours of AMD cloud compute, monthly hardware giveaways, and dedicated community support), and the AMD Embodied AI community chat. The complete demo code and challenge materials are available at \url{https://github.com/AMD-AIM/Physical_AI_Challenge}.

\section{Conclusion}

We presented four progressive technology demonstrations that together trace a complete, closed-loop path for embodied manipulation and locomotion research, built entirely on AMD's open hardware and software stack -- from Data Center and Radeon PRO GPUs to Ryzen AI edge silicon, unified by ROCm. Demo 1 showed that a manipulation policy can be trained purely in simulation and transferred directly onto a physical Franka arm, establishing the basic sim-to-real loop. Demo 2 extended that same policy family to language-grounded, semantic object selection, showing that the pipeline supports not just motor competence but instruction-following behavior in cluttered scenes. Demo 3 closed the loop in the opposite direction, reconstructing real environments via 3D Gaussian Splatting and re-injecting them into physics simulation to generate photorealistic, physically-interactive training data at scale -- turning real-world captures into an inexhaustible source of simulation-grade supervision. Demo 4 broadened the scope beyond manipulation to legged and humanoid locomotion, demonstrating that the same open compute stack scales to large-batch reinforcement learning across a diverse set of hardware, from data-center accelerators to edge-class silicon.

Taken together, these four demonstrations are not four isolated capabilities but four stages of one continuous pipeline: simulate, manipulate, reconstruct, and generalize. What makes this pipeline significant is that every stage -- 3D reconstruction, neural rendering, physics simulation, VLA policy training, reinforcement learning, and real-robot deployment -- runs on a single, open, and consistent software stack (ROCm + PyTorch) across AMD's hardware tiers, with no hand-off to a separate, CUDA-locked toolchain at any point. This end-to-end continuity is precisely what Physical AI research needs: closing the sim-to-real gap is fundamentally an iterative process, and an open, hardware-diverse, and fully interoperable compute stack lowers the barrier to that iteration for the whole community, rather than tying it to a single vendor's closed ecosystem. As the field moves from Generative AI to Agentic AI to Physical AI, we see this kind of open, end-to-end pipeline -- reproducible by any developer on the Radeon Cloud Platform -- as essential infrastructure for embodied intelligence research at scale.

\end{document}